\documentclass{sip_arXiv}
{\catcode`\ =\active\xdef {\space}}

\usepackage{amsmath} 
\usepackage{graphicx} 

\usepackage{subcaption}
\usepackage{multirow}
\usepackage{enumitem}
\usepackage[dvipsnames]{xcolor}
\usepackage[space]{grffile} 
\usepackage{tcolorbox}
\usepackage{hyperref}
\usepackage{newfloat}
\usepackage{afterpage}
\usepackage{placeins}

\newcounter{hypcount}

\newcommand{\hyp}[1]{
  \addtocounter{hypcount}{1}
  \begin{tcolorbox}
    \noindent\textbf{\emph{Hypothesis \Alph{hypcount}: }}#1
    \label{hyp-\Alph{hypcount}}
  \end{tcolorbox}
}

\newcounter{ex}
\newenvironment{example}[1][0.9]
{\vspace*{2ex}\begin{center}\begin{minipage}{#1\textwidth}}
    {\centering\refstepcounter{ex} \fontsize{9.5}{10}\selectfont\textbf{Example~\theex \label{ex:\theex} }\end{minipage}\end{center}}

\newcommand{\argmax}{\arg\!\max}

\definecolor{usc}{rgb}{0.6,0.106,0.117}
\usepackage{color}

\begin{document}%

\title[Learning from Past Mistakes]{Learning from Past Mistakes: Improving Automatic Speech Recognition Output via Noisy-Clean Phrase Context Modeling}

\author[Prashanth Gurunath Shivakumar, \textit{et al}.]{Prashanth Gurunath Shivakumar$^{1}$, Haoqi Li$^{1}$, Kevin Knight$^{2}$ and Panayiotis Georgiou $^{1}$}

\address{\add{1}{Signal Processing for Communication Understanding and Behavior Analysis Laboratory (SCUBA), University of Southern California, Los Angeles, USA}
\add{2}{Information Sciences Institute, University of Southern California, Los Angeles, USA}}

\corres{\name{Panayiotis Georgiou}
\email{georgiou@sipi.usc.edu}}

\begin{abstract}
Automatic speech recognition (ASR) systems often make unrecoverable errors due to subsystem pruning (acoustic, language and pronunciation models); for example pruning words due to acoustics using short-term context, prior to rescoring with long-term context based on linguistics.
In this work we model ASR as a phrase-based noisy transformation channel and propose an error correction system that can learn from the aggregate errors of all the independent modules constituting the ASR and attempt to invert those.
The proposed system can exploit long-term context using a neural network language model and can better choose between existing ASR output possibilities as well as re-introduce previously pruned or unseen (out-of-vocabulary) phrases.
It provides corrections under poorly performing ASR conditions without degrading any accurate transcriptions; such corrections are greater on top of out-of-domain and mismatched data ASR.
Our system consistently provides improvements over the baseline ASR, even when baseline is further optimized through recurrent neural network language model rescoring.
This demonstrates that any ASR improvements can be exploited independently and that our proposed system can potentially still provide benefits on highly optimized ASR.
Finally, we present an extensive analysis of the type of errors corrected by our system.
\end{abstract}

\keywords{Error correction, Speech recognition, Phrase-based context modeling, Noise channel estimation, Neural Network Language Model}

\maketitle

\section{Introduction}
\label{sec:intro}

Due to the complexity of human language and quality of speech signals, improving performance of automatic speech recognition (ASR) is still a challenging task. The traditional ASR comprises of three conceptually distinct modules: acoustic modeling, dictionary and language modeling.  Three modules are fairly independent of each other in research and operation.

In terms of acoustic modeling, Gaussian Mixture Model (GMM) based Hidden Markov Model (HMM) systems \cite{rabiner1989tutorial, Rabiner:1993:FSR:153687} were a standard for ASR for a long time and are still used in some of the current ASR systems. 
Lately, advances in Deep Neural Network (DNN) led to the advent of Deep Belief Networks (DBN) and Hybrid DNN-HMM 
\cite{hinton2012deep,dahl2012context}, which basically replaced the GMM with a DNN and employed a HMM for alignments. Deep Recurrent Neural Networks (RNN), particularly Long Short Term Memory (LSTM) Networks replaced the traditional DNN and DBN systems \cite{graves2013speech}. Connectionist Temporal Classification (CTC) \cite{graves2006connectionist} proved to be effective with the ability to compute the alignments implicitly under the DNN architecture, thereby eliminating the need of GMM-HMM systems for computing alignments.

The research efforts for developing efficient dictionaries or lexicons have been mainly in terms of pronunciation modeling. Pronunciation modeling was introduced to handle the intra-speaker variations \cite{strik1999modeling,wester2003pronunciation}, non-native accent variations \cite{strik1999modeling,wester2003pronunciation}, speaking rate variations found in conversational speech \cite{wester2003pronunciation} and increased pronunciation variations found in children's speech \cite{shivakumar2014improving}. Various linguistic knowledge and data-derived phonological rules were incorporated to augment the lexicon.

Research efforts in language modeling share those of the Natural Language Processing (NLP) community. By estimating the distribution of words, statistical language modeling (SLM), such as n-gram, decision tree models \cite{bahl1989tree}, linguistically motivated models \cite{moore1995combining} amount to calculating the probability distribution of different linguistic units, such as words, phrases \cite{kuo1999phrase}, sentences, and whole documents \cite{rosenfeld2000two}. Recently, Deep Neural Network based language models \cite{arisoy2012deep, mikolov2010recurrent, sundermeyer2012lstm} have also shown success in terms of both perplexity and word error rate.

Very recently, state-of-the-art ASR systems are employing end-to-end neural network models, such as sequence-to-sequence \cite{sutskever2014sequence} in an encoder-decoder architecture. The systems are trained end-to-end from acoustic features as input to predict the phonemes or characters \cite{bahdanau2016end,chan2016listen}. Such systems can be viewed as an integration of acoustic and lexicon pronunciation models. The state-of-the-art performance can be attributed towards the joint training (optimization) between the acoustic model and the lexicon models (end-to-end) enabling them to overcome the short-comings of the former independently trained models.

Several research efforts were carried out for error correction using post-processing techniques. Much of the effort involves user input used as a feedback mechanism to learn the error patterns \cite{ainsworth1992feedback,noyes1994errors}. Other work employs multi-modal signals to correct the ASR errors \cite{suhm2001multimodal, noyes1994errors}. Word co-occurrence information based error correction systems have proven quite successful \cite{sarma2004context}. In \cite{ringger1996error}, a word-based error correction technique was proposed. The technique demonstrated the ability to model the ASR as a noisy channel. In \cite{jeong2004speech}, similar technique was applied to a syllable-to-syllable channel model along with maximum entropy based language modeling.
In \cite{d2016automatic}, a phrase-based machine translation system was used to adapt a generic ASR to a domain specific grammar and vocabulary. The system trained on words and phonemes, was used to re-rank the n-best hypotheses of the ASR. In \cite{cucu2013statistical}, a phrase based machine translation system was used to adapt the models to the domain-specific data obtained by manual user-corrected transcriptions. In \cite{tam2014asr}, an RNN was trained on various text-based features to exploit long-term context for error correction. Confusion networks from the ASR have also been used for error correction. In \cite{7918446}, a bi-directional LSTM based language model was used to re-score the confusion network. In \cite{nakatani2013two}, a two step process for error correction was proposed in which words in the confusion network are re-ranked. Errors present in the confusion network are detected by conditional random fields (CRF) trained on n-gram features and subsequently long-distance context scores are used to model the long contextual information and re-rank the words in the confusion network. \cite{byambakhishig2014error,fusayasu2015word} also makes use of confusion networks along with semantic similarity information for training CRFs for error correction. 

\noindent\emph{\textbf{Our Contribution:}}
The scope of this paper is to evaluate whether subsequent transcription corrections can take place, on top of a highly optimized ASR. We hypothesize that our system can correct the errors by (i) re-scoring lattices, (ii) recovering pruned lattices, (iii) recovering unseen phrases, (iv) providing better recovery during poor recognitions, (v) providing improvements under all acoustic conditions, (vi) handling mismatched train-test conditions, (vii) exploiting longer contextual information and (viii) text regularization.
We target to satisfy the above hypotheses by proposing a Noisy-Clean Phrase Context Model (NCPCM). We introduce context of past errors of an ASR system, that consider all the automated system noisy transformations. These errors may come from any of the ASR modules or even from the noise characteristics of the signal. Using these errors we learn a noisy channel model, and apply it for error correction of the ASR output. 

\noindent Compared to the above efforts, our work differs in the following aspects:
\vspace{-\topsep}
\begin{itemize}[labelsep=5pt]

\item Error corrections take place on the output  of a state-of-the-art \emph{Large Vocabulary Continuous Speech Recognition} (LVCSR) system trained on matched data. This differs from adapting to constrained domains (e.g. \cite{cucu2013statistical,d2016automatic}) that exploit domain mismatch. This provides additional challenges both due to the larger error-correcting space (spanning larger vocabulary) and the already highly optimized ASR output.
  
\item We evaluate on a standard LVCSR task thus establishing the effectiveness, reproducibility and generalizability of the proposed correction system. This differs from past work where speech recognition was on a large-vocabulary task but subsequent error corrections were evaluated on a much smaller vocabulary.

\item  We analyze and evaluate multiple type of error corrections (including but not restricted to \emph{Out-Of-Vocabulary} (OOV) words). Most  prior work is directed towards  recovery of OOV words. 
  
\item In addition to evaluating a large-vocabulary correction system on in-domain (Fisher, 42k words) we evaluate on an out-of-domain, larger vocabulary task (TED-LIUM, 150k words), thus assessing the effectiveness of our system on challenging scenarios. In this case the adaptation is to an even bigger vocabulary, a much more challenging task to past work that only considered adaptation from large to small vocabulary tasks.
  
\item We employ multiple hypotheses of ASR to train our noisy channel model.
  
\item We employ state-of-the-art neural network based language models under the noisy-channel modeling framework which enable  exploitation of longer context.
\end{itemize}

Additionally, our proposed system comes with several advantages: (1) the system could potentially be trained without an ASR by creating a phonetic model of corruption and emulating an ASR decoder on generic text corpora, (2) the system can rapidly adapt to new linguistic patterns, e.g., can adapt to unseen words during training via contextual transformations of erroneous LVCSR outputs.

Further, our work is different from discriminative training of acoustic \cite{WOODLAND200225} models and discriminative language models (DLM) \cite{roark2007discriminative}, which are trained directly to optimize the word error rate using the reference transcripts. DLMs in particular involve optimizing, tuning, the weights of the language model with respect to the reference transcripts and are often utilized in re-ranking n-best ASR hypotheses \cite{roark2007discriminative,sagae2012hallucinated,xu2012phrasal,bikelf2012confusion,celebi2012semi}. The main distinction and advantage with our method is the NCPCM can potentially re-introduce unseen or pruned-out phrases. Our method can also operate when there is no access to lattices or n-best lists. The NCPCM can also operate on the output of a DLM system.

The rest of the paper is organized as follows: Section~\ref{sec:hyp} presents various hypotheses and discusses the different types of errors we expect to model. Section~\ref{sec:method} elaborates on the proposed technique and Section~\ref{sec:exp} describes the experimental setup and the databases employed in this work. Results and discussion are presented in Section~\ref{sec:res} and we finally conclude and present future research directions in Section~\ref{sec:conclusion}.

\section{Hypotheses}\label{sec:hyp}
In this section we analytically present cases that we hypothesize the proposed system could help with. In all of these the errors of the ASR may stem from realistic constraints of the decoding system and pruning structure, while the proposed system could exploit very long context to improve the ASR output.


Note that the vocabulary of an ASR doesn't always match the one of the error correction system. Lets consider for example, an ASR that does not have  lexicon entries for ``Prashanth'' or  ``Shivakumar'' but it has the entries ``Shiva'' and ``Kumar''. Lets also assume that this ASR consistently makes the error ``Pression'' when it hears ``Prashanth''. Given training data for the NCPCM, it  will learn the transformation ``Pression Shiva Kumar'' into ``Prashanth Shivakumar'', thus it will have a larger vocabulary than the ASR and learn to recover such errors. This demonstrates the ability to learn out-of-vocabulary entries and to rapidly adapt to new domains.

\subsection{Re-scoring Lattices}\label{hyp:rescore}

\begin{example}[0.8]
\begin{enumerate}[label*=\arabic*.]
\item\textit{``I was born in nineteen ninety three in \underline{Iraq}''}
\item\textit{``I was born in nineteen ninety three in \underline{eye rack}''}
\item\textit{``I was born in nineteen ninety three in \underline{I rack}''}
\end{enumerate}
\begin{flushleft}
Phonetic Transcription: ``\emph{ay . w aa z . b ao r n . ih n .\\ n ay n t iy n . n ay n t iy . th r iy . ih n . \underline{ay . r ae k}}''\\
\end{flushleft}
\end{example}

In Example~\ref{ex:1}, all the three samples have the same phonetic transcription.
Let us assume sample 1 is the correct transcription. Since all the three examples have the same phonetic transcription, this makes them  indistinguishable by the acoustic model. The language model is likely to down-score the sample 3. It is possible that sample 2 will score higher than sample 1 by a short context LM (e.g. bi-gram or 3-gram) i.e., ``in'' might be followed by ``eye'' more frequently than ``Iraq'' in the training corpora. This will likely result in an ASR error. Thus, although the oracle WER can be zero, the output WER is likely going to be higher due to LM choices. 
 
\hyp{An ideal error correction system can select correct options from the existing lattice.}

\subsection{Recovering Pruned Lattices}\label{hyp:pruned}
A more severe case of Example~\ref{ex:1} would be that the word ``Iraq'' was pruned out of the output lattice during decoding. This is often the case when there are memory and complexity constraints in decoding large acoustic and language models, where the decoding beam is a restricting parameter. In such cases, the word never ends up in the output lattice. Since the ASR is constrained to pick over the only existing possible paths through the decoding lattice, an error is inevitable in the final output.

\hyp{An ideal error correction system can generate words or phrases that were erroneously pruned during the decoding process.}

\subsection{Recovery of Unseen Phrases}\label{hyp:recover}
On the other hand, an extreme case of Example~\ref{ex:1} would be that the word ``Iraq'' was never seen in the training data (or is out-of-vocabulary), thereby not appearing in the ASR lattice. This would mean the ASR is forced to select among the other hypotheses even with a low confidence (or output an unknown, $<unk>$, symbol) resulting in a similar error as before. This is often the case due to the constant evolution of human language or in the case of a new domain. For example, names such as ``Al Qaeda'' or ``ISIS'' were non-existent in our vocabularies a few years ago.

\hyp{An ideal error correction system can generate words or phrases that are out of vocabulary (OOV) and thus not in the ASR output.}

\subsection{Better Recovery during Poor Recognitions}\label{hyp:low_conditions}
An ideal error correction system would provide more improvements for poor recognitions from an ASR. Such a system could potentially offset for the ASR's low performance providing consistent performance over varying audio and recognition conditions. In real-life conditions, the ASR often has to deal with varying level of ``mismatched train-test'' conditions, where relatively poor recognition results are commonplace.
 
\hyp{An ideal error correction system can provide more corrections when the ASR performs poorly, thereby offsetting ASR's performance drop (e.g.\ during mismatched train-test conditions).}

\subsection{Improvements under all Acoustic Conditions}\label{hyp:wer}
An error correction system which performs well during tough recognition conditions, as per \emph{Hypothesis}~\ref{hyp-D} is no good if it degrades good recognizer output. Thus, in addition to our \emph{Hypothesis}~\ref{hyp-D}, an ideal system would cause no degradation on good ASR output.
Such a system can be hypothesized to consistently improve upon and provide benefits over any ASR system including state-of-the-art recognition systems. An ideal system would provide improvements over the entire spectrum of ASR performance (WER).

\hyp{An ideal error correction system can not only provide improvements during poor recognitions, but also preserves good speech recognition.}

\subsection{Adaptation}\label{hyp:adapt}
We hypothesize that the proposed system would help in adaptation over mismatched conditions. The mismatch could manifest in terms of acoustic conditions and lexical constructs. The adaptation can be seen as a consequence of \emph{Hypothesis}~\ref{hyp-D} \&~\ref{hyp-E}. In addition, the proposed model is capable of capturing patterns of language use manifesting in specific speaker(s) and domain(s). Such a system could eliminate the need of retraining the ASR model for mismatched environments.

\hyp{An ideal error correction system can aid in mismatched train-test conditions.}

\subsection{Exploit Longer Context}\label{hyp:long_context}
\begin{example}[0.9]
\begin{itemize}
\item\textit{``\textbf{Eyes} melted, when he placed \underline{his hand on her shoulders.}''}
\item\textit{``\textbf{Ice} melted, when he placed \underline{it on the table.}''}
\end{itemize}
\end{example}

The complex construct of human language and understanding enables recovery of lost or corrupted information over different temporal resolutions. For instance, in the above Example~\ref{ex:2}, both the phrases, ``Eyes melted, when he placed'' and ``Ice melted, when he placed'' are valid when viewed within its shorter context and have identical phonetic transcriptions. The succeeding phrases, underlined, help in discerning whether the first word is ``Eyes'' or ``Ice''. We hypothesize that an error correction model capable of utilizing such longer contexts is beneficial. As new models for phrase based mapping, such as sequence to sequence models \cite{sutskever2014sequence}, become applicable this becomes even more possible and desirable.

\hyp{An ideal error correction system can exploit longer context than the ASR for better corrections.}

\subsection{Regularization}\label{hyp:reqularization}
\begin{example}[0.7]
\begin{enumerate}[label*=\arabic*.] \item
\begin{itemize}
\item\textit{``I guess \underline{'cause} I went on a I went on a ...''}
\item\textit{``I guess \underline{because} I went on a I went on a ...''}
\end{itemize}\item
\begin{itemize}
\item\textit{``i was born in nineteen ninety two''}
\item\textit{``i was born in 1992''}
\end{itemize}\item
\begin{itemize}
\item\textit{``i was born on nineteen twelve''}
\item\textit{``i was born on 19/12''}
\end{itemize}
\end{enumerate}
\end{example}

As per the 3 cases shown in Example~\ref{ex:3}, although both the hypotheses for each of them are correct, there are some irregularities present in the language syntax. Normalization of such surface form representation can increase readability and usability of output. Unlike traditional ASR, where there is a need to explicitly program such regularizations, our system is expected to learn, given appropriate training data, and incorporate regularization into the model.

\hyp{An ideal error correction system can be deployed as an automated text regularizer.}

\section{Methodology}
\label{sec:method}

\begin{figure*}[b]
\begin{center}
\includegraphics[width=\textwidth]{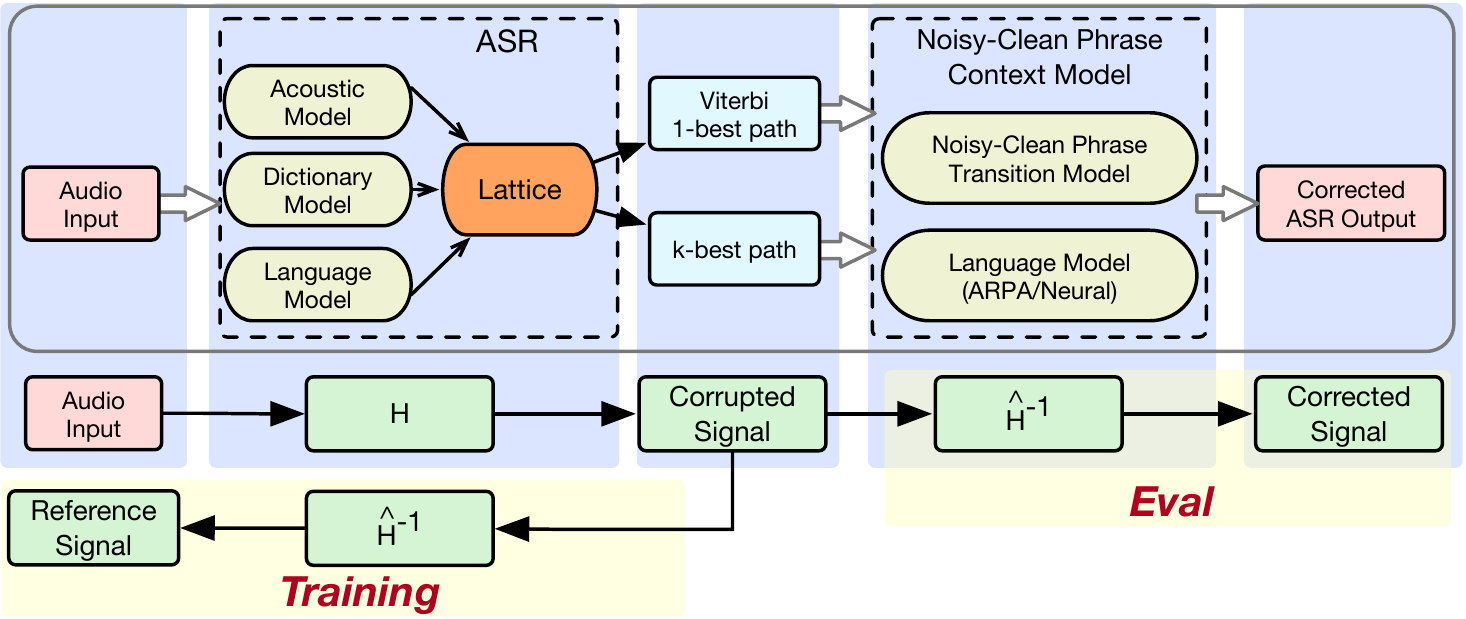}
\caption{Overview of NCPCM}
\label{fig:overview}
\end{center}
\end{figure*}

The overview of the proposed model is shown in Figure~\ref{fig:overview}. In our paper, the ASR is viewed as a noisy channel (with transfer function $H$), and we learn a model of this channel, $\widehat{H}^{-1}$ (estimate of inverse transfer function $H^{-1}$) by using the corrupted ASR outputs (equivalent to signal corrupted by $H$) and their reference transcripts. Later on, we use this model to correct the errors of the ASR.

The noisy channel modeling mainly can be divided into word-based and phrase-based channel modeling. We will first introduce previous related work, and 
then our proposed NCPCM.

\subsection{Previous related work}
\subsubsection{Word-based Noisy Channel Modeling}\label{sec:word_based}
In \cite{ringger1996error}, the authors adopt word-based noisy channel model borrowing ideas from a word-based statistical machine translation developed by IBM \cite{brown1990statistical}. It is used as a post-processor module to correct the mistakes made by the ASR. The word-based noisy channel modeling can be presented as:
\begin{align*}
\hat{W} & = \argmax_{W_{\text{clean}}} P(W_{\text{clean}}|W_{\text{noisy}}) \\
& = \argmax_{W_{\text{clean}}} P(W_{\text{noisy}}|W_{\text{clean}}) P_{\text{LM}}(W_{\text{clean}})
\end{align*}
where $\hat{W}$ is the corrected output word sequence, $P(W_{\text{clean}}|W_{\text{noisy}})$ is the posterior probability, $P(W_{\text{noisy}}|W_{\text{clean}})$ is the channel model and $P_{\text{LM}}(W_{\text{clean}})$ is the language model. In \cite{ringger1996error}, authors hypothesized that introducing many-to-one and one-to-many word-based channel modeling (referred to as fertility model) could be more effective, but was not implemented in their work.

\subsubsection{Phrase-based Noisy Channel Modeling}
Phrase-based systems were introduced in application to phrase-based statistical translation system \cite{koehn2003statistical} and were shown to be superior to the word-based systems. Phrase based transformations are similar to word-based models with the exception that the fundamental unit of observation and transformation is a phrase (one or more words). It can be viewed as a super-set of the word-based  \cite{brown1990statistical} and the fertility \cite{ringger1996error} modeling systems.

\subsection{Noisy-Clean Phrase Context Modeling}
We extend the ideas by proposing a complete phrase-based channel modeling for error correction which incorporates the many-to-one and one-to-many as well as many-to-many words (phrase) channel modeling for error-correction. This also allows the model to better capture errors of varying resolutions made by the ASR. As an extension, it uses a distortion modeling to capture any re-ordering of phrases during error-correction. Even though we do not expect big benefits from the distortion model (i.e., the order of the ASR output is usually in agreement with the audio representation), we include it in our study for examination. It also uses a word penalty to control the length of the output. The phrase-based noisy channel modeling can be represented as:
\begin{align}
\hat{p} & = \argmax_{p_{\text{clean}}} P(p_{\text{clean}}|p_{\text{noisy}}) \label{eq:ncpm_decoding}\\
& = \argmax_{p_{\text{clean}}} P(p_{\text{noisy}}|p_{\text{clean}}) P_{\text{LM}}(p_{\text{clean}}) w_{\text{length}}(p_{\text{clean}}) \nonumber 
\end{align}
where $\hat{p}$ is the corrected sentence, $p_{\text{clean}}$ and $p_{\text{noisy}}$ are the reference and noisy sentence respectively. $w_{\text{length}}(p_{\text{clean}})$ is the output word sequence length penalty, used to control the output sentence length, and $P(p_{\text{noisy}}|p_{\text{clean}})$ is decomposed into:
\begin{align}
P(p_{\text{noisy}}^{I}|p_{\text{clean}}^{I}) = \prod_{i=1}^{I} \phi(p_{\text{noisy}}^{i}|p_{\text{clean}}^{i}) D(start_i - end_{i-1}) \label{eq:ncpm}
\end{align}
where $\phi(p_{\text{noisy}}^{i}|p_{\text{clean}}^{i})$ is the phrase channel model or phrase translation table, $p_{\text{noisy}}^{I}$ and $p_{\text{clean}}^{I}$ are the sequences of $I$ phrases in noisy and reference sentences respectively and $i$ refers to the $i^{th}$ phrase in the sequence. $D(start_i - end_{i-1})$ is the distortion model. $start_i$ is the start position of the noisy phrase that was corrected to the $i^{th}$ clean phrase, and $end_{i-1}$ is the end position of the noisy phrase corrected to be the ${i-1}^{th}$ clean phrase. 

\subsection{Our Other Enhancements}
In order to effectively demonstrate our idea, we employ (i) neural language models, to introduce long term context and justify that the longer contextual information is beneficial for error corrections; (ii) minimum error rate training (MERT) to tune and optimize the model parameters using development data.

\subsubsection{Neural Language Models}
Neural network based language models have been shown to be able to model higher order n-grams more efficiently \cite{arisoy2012deep, mikolov2010recurrent, sundermeyer2012lstm}. 
In \cite{jeong2004speech}, a more efficient language modeling using maximum entropy was shown to help in noisy-channel modeling of a syllable-based ASR error correction system. 

Incorporating such language models would aid the error-correction by exploiting the longer context information. Hence, we adopt two types of neural network language models in this work. (i) Feed-forward neural network which is trained using a sequence of one-hot word representation along with the specified context \cite{vaswani2013decoding}. (ii) Neural network joint model (NNJM) language model \cite{devlin2014fast}. This is trained in a similar way as in (i), but the context is augmented with noisy ASR observations with a specified context window. 
Both the models employed are feed-forward neural networks since they can be incorporated directly into the noisy channel modeling.
The recurrent neural network LM could potentially be used during phrase-based decoding by employing certain caching and approximation tricks \cite{alkhouli2015investigations}.
Noise Contrastive Estimation was used to handle the large vocabulary size output.
\subsubsection{Minimum Error Rate Training (MERT)}
One of the downsides of the noisy channel modeling is that the model is trained to maximize the likelihood of the seen data and there is no direct optimization to the end criteria of WER. MERT optimizes the model parameters (in our case weights for language, phrase, length and distortion models) with respect to the desired end evaluation criterion. MERT was first introduced in application to statistical machine translation providing significantly better results \cite{och2003minimum}. We apply MERT to tune the model on a small set of development data.

\begin{table*}[b]
\centering
\begin{tabular}{|c|c|c|c|c|c|c|c|c|c|}
\hline
\multirow{2}{*}{Database} & \multicolumn{3}{c|}{Train} & \multicolumn{3}{c|}{Development} & \multicolumn{3}{c|}{Test} \\ 
& Hours & Utterances & Words & Hours & Utterances & Words & Hours & Utterances & Words \\
\hline
Fisher English & 1,890.5 & 1,833,088 & 20,724,957 & 4.7 & 4906 & 50,245 & 4.7 & 4914 & 51,230 \\ \hline
TED-LIUM & - & - & - & 1.6 & 507 & 17,792 & 2.6 & 1155 & 27,512 \\
\hline
\end{tabular}
\caption{Database split and statistics}\label{tab:data_split}
\label{tab:result_table_accept}
\end{table*}

\section{Experimental Setup}\label{sec:exp}

\subsection{Database}
For training, development, and evaluation, we employ Fisher English Training Part 1, Speech (LDC2004S13) and Fisher English Training Part 2, Speech (LDC2005S13) corpora \cite{cieri2004fisher}. The Fisher English Training Part 1, is a collection of conversation telephone speech with 5850 speech samples of up to 10 minutes, approximately 900 hours of speech data. The Fisher English Training Part 2, contains an addition of 5849 speech samples, approximately 900 hours of telephone conversational speech. The corpora is split into training, development and test sets for experimental purposes as shown in Table \ref{tab:data_split}. The splits of the data-sets are consistent over both the ASR and the subsequent noisy-clean phrase context model. The development dataset was used for tuning the phrase-based system using MERT.

We also test the system under mismatched training-usage conditions on TED-LIUM. TED-LIUM is a dedicated ASR corpus consisting of 207 hours of TED talks \cite{rousseau2014enhancing}. The data-set was chosen as it is significantly different to Fisher Corpus. Mismatch conditions include: (i) variations in channel characteristics, Fisher, being a telephone conversations corpus, is sampled at 8kHz where-as the TED-LIUM is originally 16kHz, (ii) noise conditions, the Fisher recordings are significantly noisier, (iii) utterance lengths, TED-LIUM has longer conversations since they are extracted from TED talks, (iv) lexicon sizes, vocabulary size of TED-LIUM is much larger with 150,000 words where-as Fisher has 42,150 unique words, (v) speaking intonation, Fisher being telephone conversations is spontaneous speech, whereas the TED talks are more organized and well articulated. Factors (i) and (ii) mostly affect the performance of ASR due to acoustic differences while (iii) and (iv) affect the language aspects, (v) affects both the acoustic and linguistic aspects of the ASR.

\subsection{System Setup}
\subsubsection{Automatic Speech Recognition System}
We used the Kaldi Speech Recognition Toolkit \cite{povey2011kaldi} to train the ASR system. In this paper, the acoustic model was trained as a DNN-HMM hybrid system. A tri-gram maximum likelihood estimation (MLE) language model was trained on the transcripts of the training dataset. The CMU pronunciation dictionary \cite{weide1998cmu} was adopted as the lexicon.  The resulting ASR is state-of-the-art both in architecture and performance and as such additional gains on top of this ASR are challenging.
\subsubsection{Pre-processing}
The reference outputs of ASR corpus contain non-verbal signs, such as [laughter], [noise] etc. These event signs might corrupt the phrase context model since there is little contextual information between them. Thus, in this paper, we cleaned our data by removing all these non-verbal signs from dataset. 
The text data is subjected to traditional tokenization to handle special symbols.
Also, to prevent data sparsity issues, we restricted all of the sample sequences to a maximum length of 100 tokens (given that the database consisted of only 3 sentences having more than the limit).
The NCPCM has two distinct vocabularies, one associated with the ASR transcripts and the other one pertaining to the ground-truth transcripts.
The ASR dictionary is often smaller than the ground-truth transcript mainly because of not having a pronunciation-phonetic transcriptions for certain words, which usually is the case for names, proper-nouns, out-of-language words, broken words etc.

\subsubsection{NCPCM}
We use the Moses toolkit \cite{koehn2007moses} for phrase based noisy channel modeling and MERT optimization. 
The first step in the training process of NCPCM is the estimation of the word alignments. IBM models are used to obtain the word alignments in both the directions (reference-ASR and ASR-reference). 
The final alignments are obtained using heuristics (starting with the intersection of the two alignments and then adding the additional alignment points from the union of two alignments). 
For computing the alignments ``mgiza'', a multi-threaded version of GIZA++ toolkit \cite{och03:asc} was employed.
Once the alignments are obtained, the lexical translation table is estimated in the maximum likelihood sense.
Then on, all the possible phrases along with their word alignments are generated . 
A max phrase length of 7 was set for this work.
The generated phrases are scored to obtain a phrase translation table with estimates of phrase translation probabilities.
Along with the phrase translation probabilities, word penalty scores (to control the translation length) and re-ordering/distortion costs (to account for possible re-ordering) are estimated.
Finally, the NCPCM model is obtained as in the equation~\ref{eq:ncpm}.
During decoding equation~\ref{eq:ncpm_decoding} is utilized.

For training the MLE n-gram models, SRILM toolkit \cite{stolcke2002srilm} was adopted. 
Further we employ the Neural Probabilistic Language Model Toolkit \cite{vaswani2013decoding} to train the neural language models.
The neural network was trained for 10 epochs with an input embedding dimension of 150 and output embedding dimension of 750, with a single hidden layer.
The weighted average of all input embeddings was computed for padding the lower-order estimates as suggested in \cite{vaswani2013decoding}.

The NCPCM is an ensemble of phrase translation model, language model, translation length penalty, re-ordering models.
Thus the tuning of the weights associated with each model is crucial in the case of proposed phrase based model.
We adopt the line-search based method of MERT \cite{bertoldi2009improved}.
We try two optimization criteria with MERT, i.e., using BLEU(B) and WER(W).

\subsection{Baseline Systems}
We adopt four different baseline systems because of their relevance to this work:\\
\textbf{Baseline-1:} \emph{ASR Output}: The raw performance of the ASR system, because of its relevance to the application of the proposed model.\\
\textbf{Baseline-2:} \emph{Re-scoring lattices using RNN-LM}: In order to evaluate the performance of the system with more recent re-scoring techniques, we train a recurrent-neural network with an embedding dimension of 400 and sigmoid activation units. 
Noise contrastive estimation is used for training the network and is optimized on the development data set which is used as a stop criterion.
Faster-RNNLM \footnote{https://github.com/yandex/faster-rnnlm} toolkit is used to train the recurrent-neural network.
For re-scoring, 1000-best ASR hypotheses are decoded and the old LM (MLE) scores are removed.
The RNN-LM scores are computed from the trained model and interpolated with the old LM.
Finally, the 1000-best hypotheses are re-constructed into lattices, scored with new interpolated LM and decoded to get the new best path hypothesis.
\textbf{Baseline-3:} \emph{Word-based noisy channel model}: In order to compare to a prior work described in Section~\ref{sec:word_based} which is based on \cite{ringger1996error}.
The word-based noisy channel model is created in a similar way as the NCPCM model with three specific exceptions: (i) the max-phrase length is set to 1, which essentially converts the phrase based model into word based, (ii) a bi-gram LM is used instead of a tri-gram or neural language model, as suggested in \cite{ringger1996error}, (iii) no re-ordering/distortion model and word penalties are used.\\
\textbf{Baseline-4:} \emph{Discriminative Language Modeling (DLM)}: Similar to the proposed work, DLM makes use of the reference transcripts to tune language model weights based on specified feature sets in order to re-rank the n-best hypothesis. Specifically, we employ the perceptron algorithm \cite{roark2007discriminative} for training DLMs. 
The baseline system is trained using unigrams, bigrams and trigrams (as in \cite{bikelf2012confusion,xu2012phrasal,sagae2012hallucinated}) for a fair comparison with the proposed NCPCM model.
We also provide results with an extended feature set comprising of rank-based features and ASR LM and AM scores.
Refr (Reranker framework) is used for training the DLMs \cite{bikel2013refr} following most recommendations from \cite{bikelf2012confusion}. 
100-best ASR hypotheses are used for training and re-ranking purposes.

\subsection{Evaluation Criteria}
The final goal of our work is to show improvements in terms of the transcription accuracy of the overall system. Thus, we provide word error rate as it is a standard in the ASR community. Moreover, Bilingual Evaluation Understudy (BLEU) score \cite{papineni2002bleu} is used for evaluating our work, since our model can be also treated as a transfer-function (``translation'') system from ASR output to NCPCM output.  

\afterpage{\FloatBarrier}
\begin{table*}[p]
		\vspace{25mm}
		\resizebox{0.97\textwidth}{!}{\begin{minipage}{1.1\textwidth}\begin{center}
    \begin{tabular}{|p{1mm}p{0.99\textwidth}|}
        \hline
				1.&\begin{enumerate}[label=\alph *),nosep,align=left,labelwidth=20mm,leftmargin=\dimexpr\labelwidth+\labelsep\relax]
        \item[\textbf{REF:}] \colorbox{SpringGreen}{oysters} clams and mushrooms i think
        \item[\textbf{ASR:}] \colorbox{Dandelion}{wasters} clams and mushrooms they think
				\item[\textbf{ORACLE:}] \colorbox{Dandelion}{wasters} clams and mushrooms i think
				\item[\textbf{NCPCM:}] \colorbox{SpringGreen}{oysters} clams and mushrooms they think
        \item[] \hfill {\color{red} Example of hypotheses B}
				\end{enumerate}\\
        \hline
				2.&\begin{enumerate}[label=\alph *),nosep,align=left,labelwidth=20mm,leftmargin=\dimexpr\labelwidth+\labelsep\relax]
        \item[\textbf{REF:}] yeah we had this awful month this winter where it was like a good day if it got up to thirty it was \colorbox{SpringGreen}{ridiculously} cold
        \item[\textbf{ASR:}] yeah we had this awful month uh this winter where it was like a good day if i got up to thirty was \colorbox{Dandelion}{ridiculous lee} cold
        \item[\textbf{ORACLE:}] yeah we had this awful month this winter where it was like a good day if it got up to thirty it was \colorbox{Dandelion}{ridiculous} the cold
				\item[\textbf{NCPCM:}] yeah we had this awful month uh this winter where it was like a good day if i got up to thirty it was \colorbox{SpringGreen}{ridiculously} cold
        \item[] \hfill {\color{red} Example of hypotheses A, B, G}
				\end{enumerate}\\
				\hline
				3.&\begin{enumerate}[label=\alph *),nosep,align=left,labelwidth=20mm,leftmargin=\dimexpr\labelwidth+\labelsep\relax]
        \item[\textbf{REF:}] oh well it depends on whether you agree that \colorbox{SpringGreen}{al qaeda} came right out of afghanistan
				\item[\textbf{ASR:}] oh well it depends on whether you agree that \colorbox{Dandelion}{al $<$unk$>$} to came right out of afghanistan
				\item[\textbf{ORACLE:}] oh well it depends on whether you agree that \colorbox{Dandelion}{al $<$unk$>$} to came right out of afghanistan
				\item[\textbf{NCPCM:}] oh well it depends on whether you agree that \colorbox{SpringGreen}{al qaeda} to came right out of afghanistan
        \item[] \hfill {\color{red} Example of hypotheses C}
				\end{enumerate}\\
        \hline
				4.&\begin{enumerate}[label=\alph *),nosep,align=left,labelwidth=20mm,leftmargin=\dimexpr\labelwidth+\labelsep\relax]
        \item[\textbf{REF:}] they \colorbox{SpringGreen}{laugh} because everybody else is laughing and not because it's really funny
				\item[\textbf{ASR:}] they \colorbox{Dandelion}{laughed} because everybody else is laughing and not because it's really funny
				\item[\textbf{ORACLE:}] they \colorbox{SpringGreen}{laugh} because everybody else is laughing and not because it's really funny
				\item[\textbf{NCPCM:}] they \colorbox{SpringGreen}{laugh} because everybody else is laughing and not because it's really funny
        \item[] \hfill {\color{red} Example of hypotheses A, G}
				\end{enumerate}\\
				\hline
				5.&\begin{enumerate}[label=\alph *),nosep,align=left,labelwidth=20mm,leftmargin=\dimexpr\labelwidth+\labelsep\relax]
				\item[\textbf{REF:}] yeah \colorbox{SpringGreen}{especially} like if you go out for ice cream or something
				\item[\textbf{ASR:}] yeah \colorbox{Dandelion}{it specially} like if you go out for ice cream or something
				\item[\textbf{ORACLE:}] yeah \colorbox{Dandelion}{it's} \colorbox{SpringGreen}{especially} like if you go out for ice cream or something
				\item[\textbf{NCPCM:}] yeah \colorbox{SpringGreen}{especially} like if you go out for ice cream or something
        \item[] \hfill {\color{red} Example of hypotheses A}
				\end{enumerate}\\
        \hline
				6.&\begin{enumerate}[label=\alph *),nosep,align=left,labelwidth=20mm,leftmargin=\dimexpr\labelwidth+\labelsep\relax]
				\item[\textbf{REF:}] we don't have a lot of that around we \colorbox{SpringGreen}{kind of} live in a nicer area
        \item[\textbf{ASR:}] we don't have a lot of that around we \colorbox{Dandelion}{kinda} live in a nicer area
				\item[\textbf{ORACLE:}] we don't have a lot of that around we \colorbox{SpringGreen}{kind of} live in a nicer area
        \item[\textbf{NCPCM:}] we don't have a lot of that around we \colorbox{SpringGreen}{kind of} live in a nicer area
        \item[] \hfill {\color{red} Example of hypotheses A, H}
				\end{enumerate}\\
        \hline
    \end{tabular}
		\end{center}\end{minipage}}
		\captionsetup{justification=centering}
		\vspace{2mm}
		\caption{\small{Analysis of selected sentences.\\
		REF: Reference ground-truth transcripts; ASR: Output ASR transcriptions;\\
		ORACLE: Best path through output lattice given the ground-truth transcript; NCPCM: Transcripts after NCPCM error-correction\\
		Green color highlights correct phrases. Orange color highlights incorrect phrases.}}
    \label{tab:analysis}
\end{table*}
\section{Results and Discussion}\label{sec:res}

In this section we demonstrate the ability of our proposed NCPCM in validating our hypotheses A-H from Section~\ref{sec:hyp} along with the experimental results. The experimental results are presented in three different tasks: (i) overall WER experiments, highlighting the improvements of the proposed system, presented in Tables~\ref{tab:results},~\ref{tab:ood} \& ~\ref{tab:lm_results}, (ii) detailed analysis of WERs over subsets of data, presented in Figures~\ref{fig:wer_analysis} \& ~\ref{fig:domain_split}, and (iii) analysis of the error corrections, presented in Table~\ref{tab:analysis}. The assessment and discussions of each task is structured similar to Section~\ref{sec:hyp} to support their respective claims.

\subsection{Re-scoring Lattices}\label{res:rescore}
Table~\ref{tab:analysis} shows selected samples through the process of the proposed error correction system. In addition to the reference, ASR output and the proposed system output, we provide the ORACLE transcripts to assess the presence of the correct phrase in the lattice. Cases 4-6 from Table~\ref{tab:analysis} have the correct phrase in the lattice, but get down-scored in the ASR final output which is then recovered by our system as hypothesized in ~\emph{Hypothesis}~\ref{hyp-A}.

\subsection{Recovering Pruned Lattices}\label{res:pruned}
In the cases 1 and 2 from Table~\ref{tab:analysis}, we see the correct phrases are not present in the ASR lattice, although they were seen in the training and are present in the vocabulary. However, the proposed system manages to recover the phrases as discussed in \emph{Hypothesis}~\ref{hyp-B}. Moreover, Case 2 also demonstrates an instance where the confusion occurs due to same phonetic transcriptions (``ridiculously'' versus ``ridiculous lee'') again supporting \emph{Hypothesis}~\ref{hyp-A}.

\subsection{Recovery of Unseen Phrases}\label{res:recover}
Case 3 of Table~\ref{tab:analysis}, demonstrates an instance where the word ``qaeda'' is absent from the ASR lexicon (vocabulary) and hence absent in the decoding lattice. This forces the ASR to output an unknown-word token ($<unk>$). We see that the system recovers an out-of-vocabulary word ``qaeda'' successfully as claimed in ~\emph{Hypothesis}~\ref{hyp-C}.

\begin{figure*}[t] 
  \begin{subfigure}[t]{0.5\textwidth}
	  \centering
		\includegraphics[width=\textwidth]{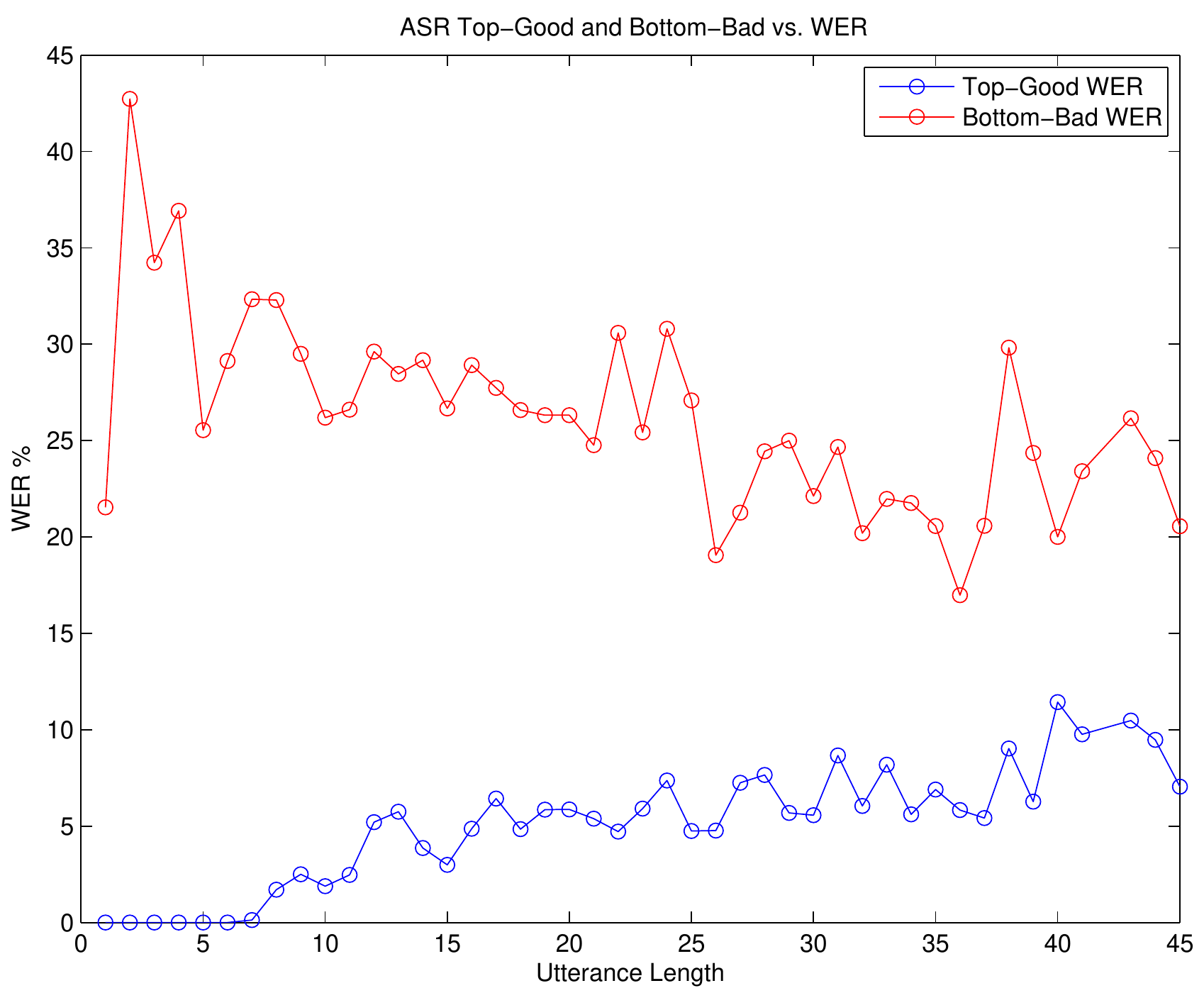}
		\caption{In Domain (Fisher)}
		\label{fig:in_domain}
	\end{subfigure}
  \begin{subfigure}[t]{0.5\textwidth}
    \centering
		\includegraphics[width=\textwidth]{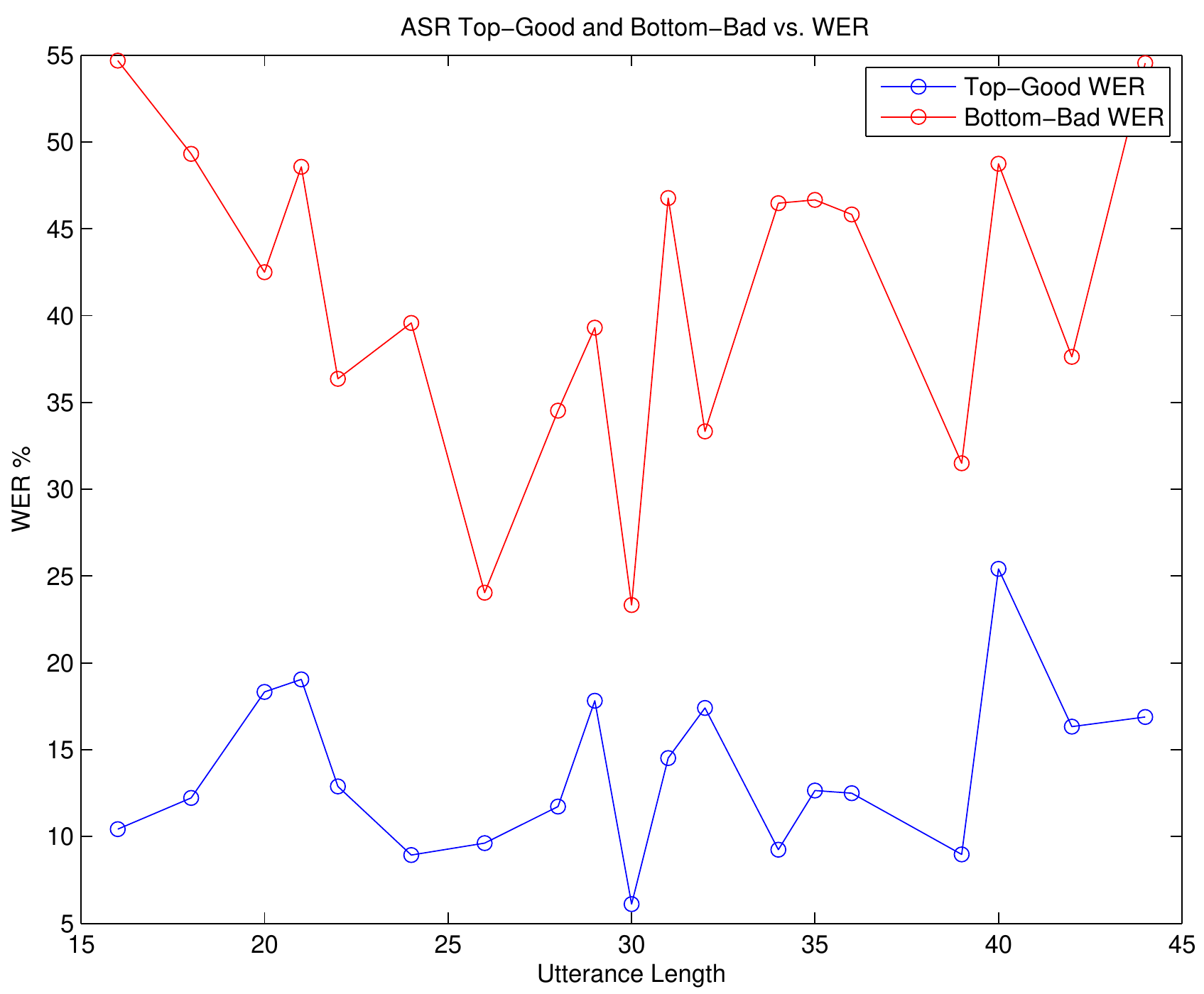}
		\caption{Out of Domain (TEDLIUM)}
		\label{fig:out_of_domain}
	\end{subfigure}
	\caption{Top-Good, Bottom-Bad WER Splits. As we can see the WER for top-good is often 0\%, which leaves no margin for improvement. We will see the impact of this later, as in Fig.~\ref{fig:wer_analysis}}\label{fig:domain_split}
\end{figure*}

\begin{figure*}
\captionsetup{justification=centering}
  \begin{subfigure}[t]{0.47\textwidth}
	  \centering
		\includegraphics[width=\textwidth]{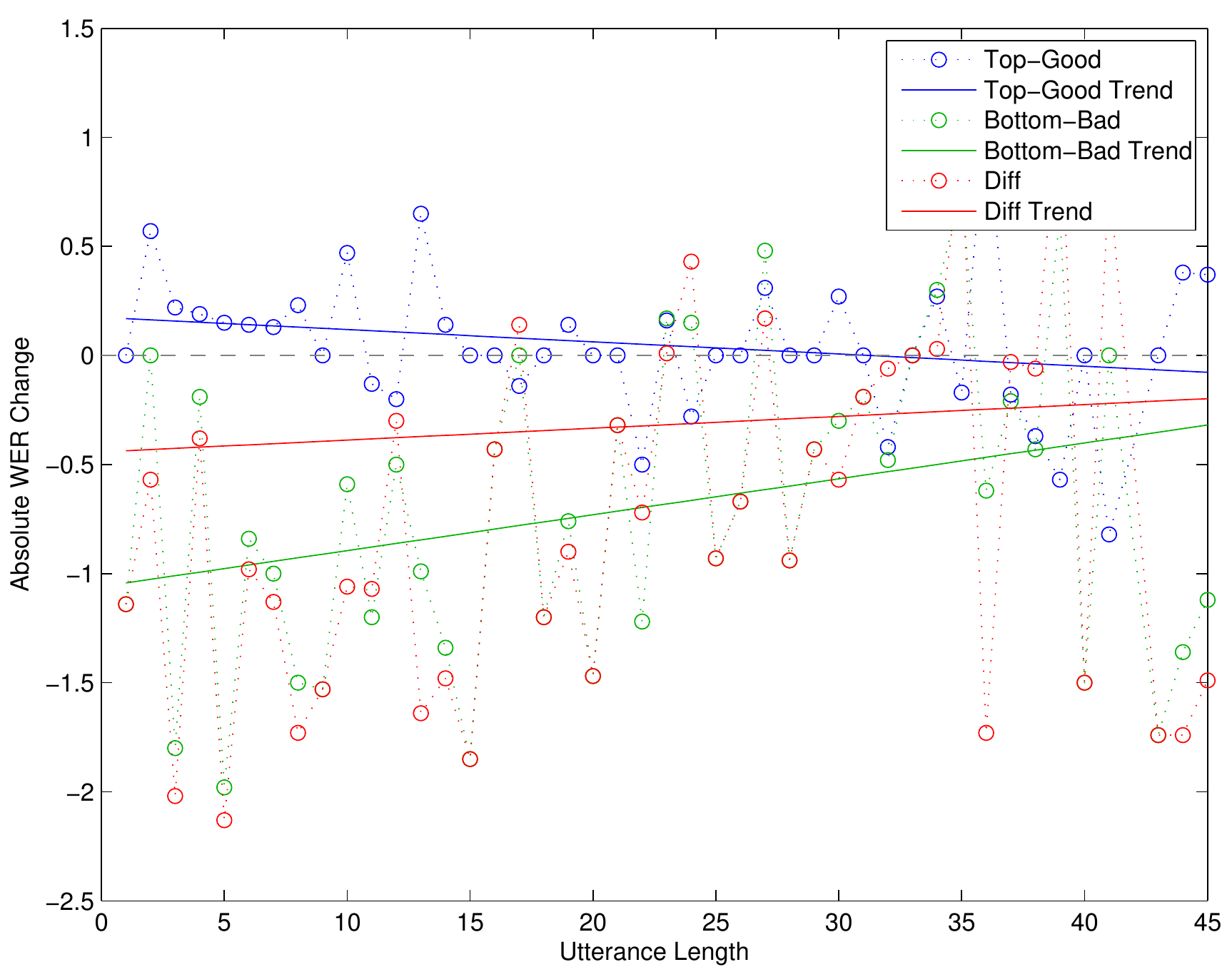}
    \caption{Dev: NCPCM + MERT(W)}
		\label{fig:1best_dev}
	\end{subfigure}
  \hfill
	\begin{subfigure}[t]{0.47\textwidth}
		\centering
		\includegraphics[width=\textwidth]{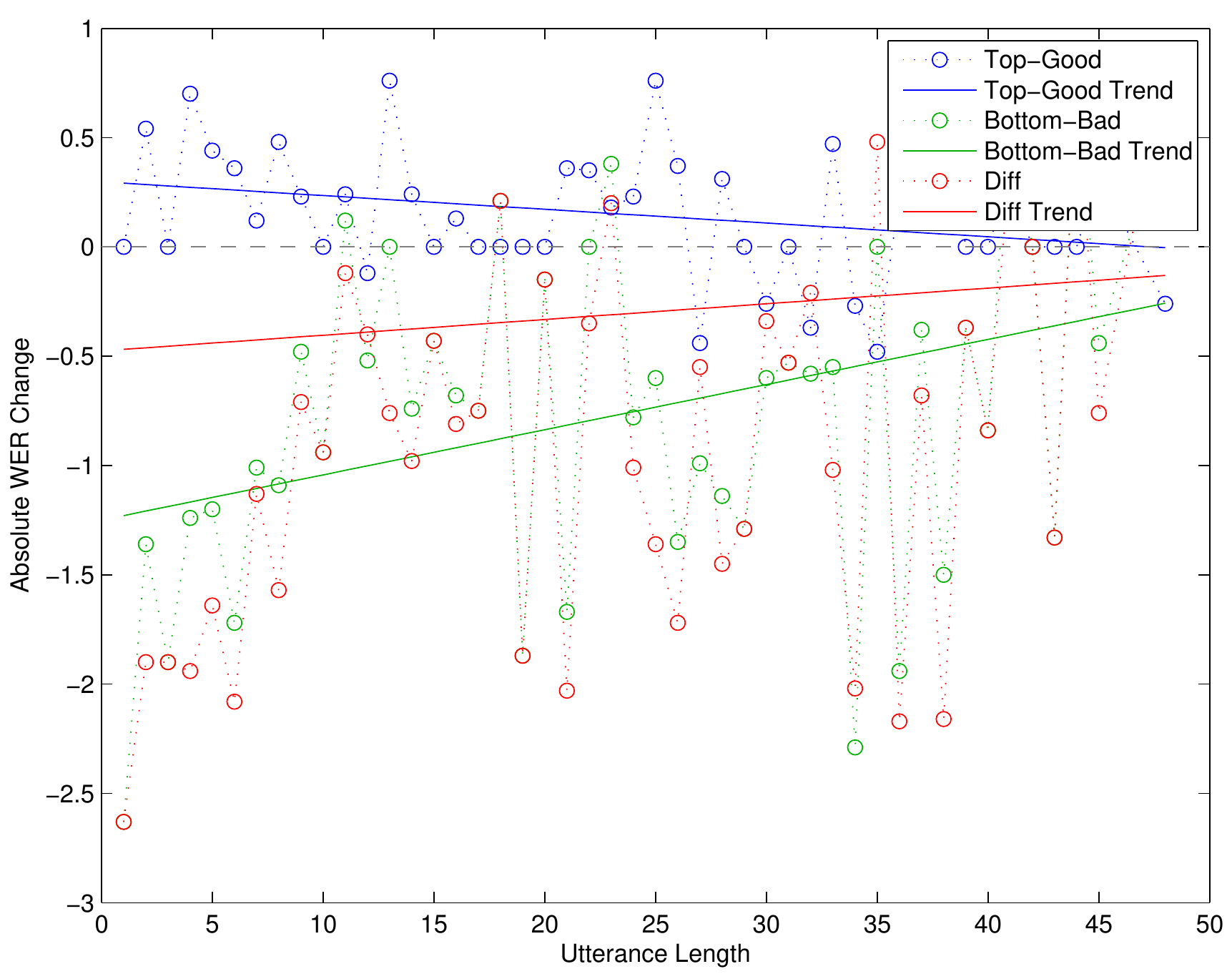}
    \caption{Test: NCPCM + MERT(W)}
		\label{fig:1best_test}
	\end{subfigure}
	\begin{subfigure}[t]{0.47\textwidth}
	  \centering
		\includegraphics[width=\textwidth]{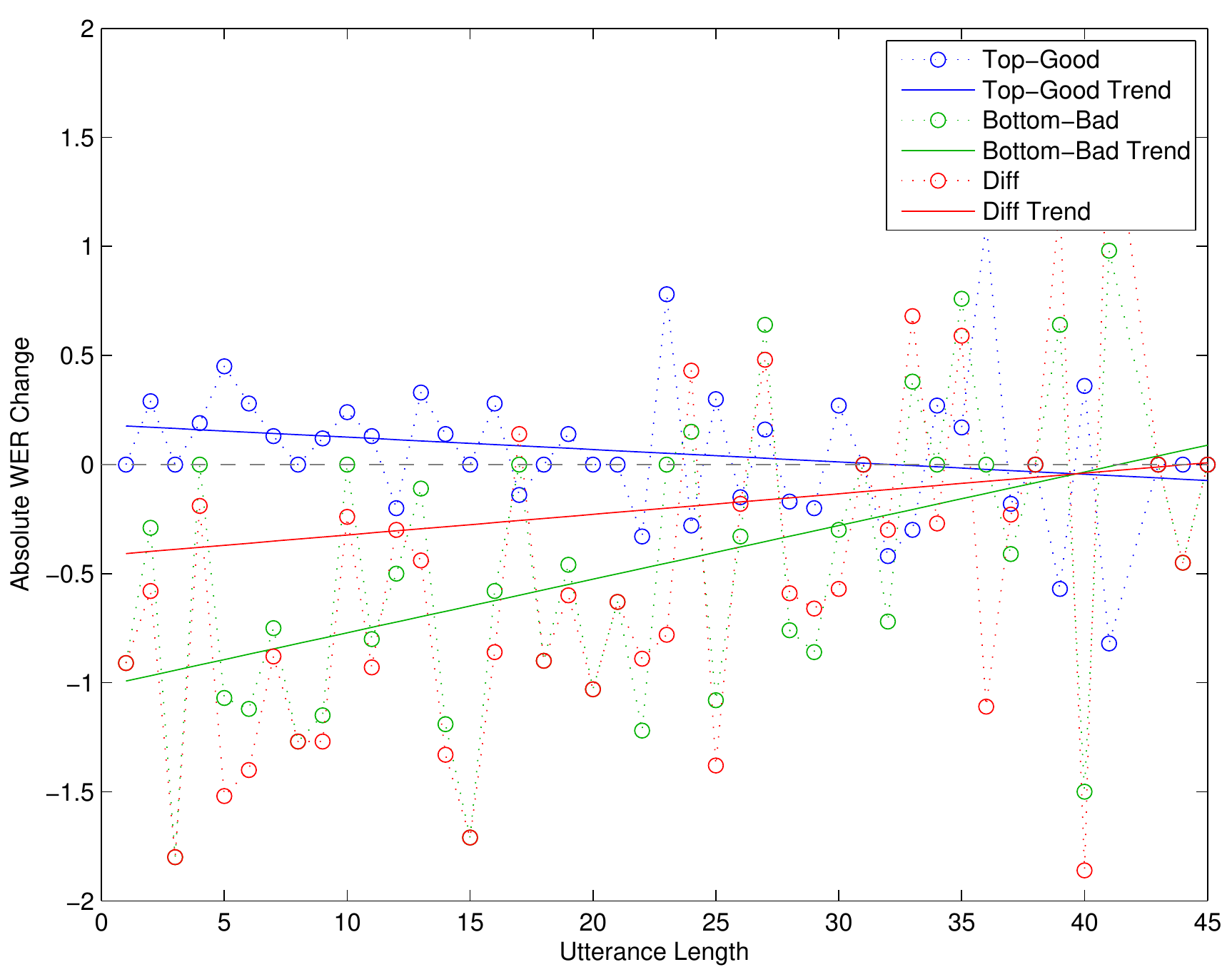}
    \caption{Dev: NCPCM + 5gram NNLM + MERT(W)}
		\label{fig:5gNNLM_dev}
	\end{subfigure}
  \hfill
	\begin{subfigure}[t]{0.47\textwidth}
		\centering
		\includegraphics[width=\textwidth]{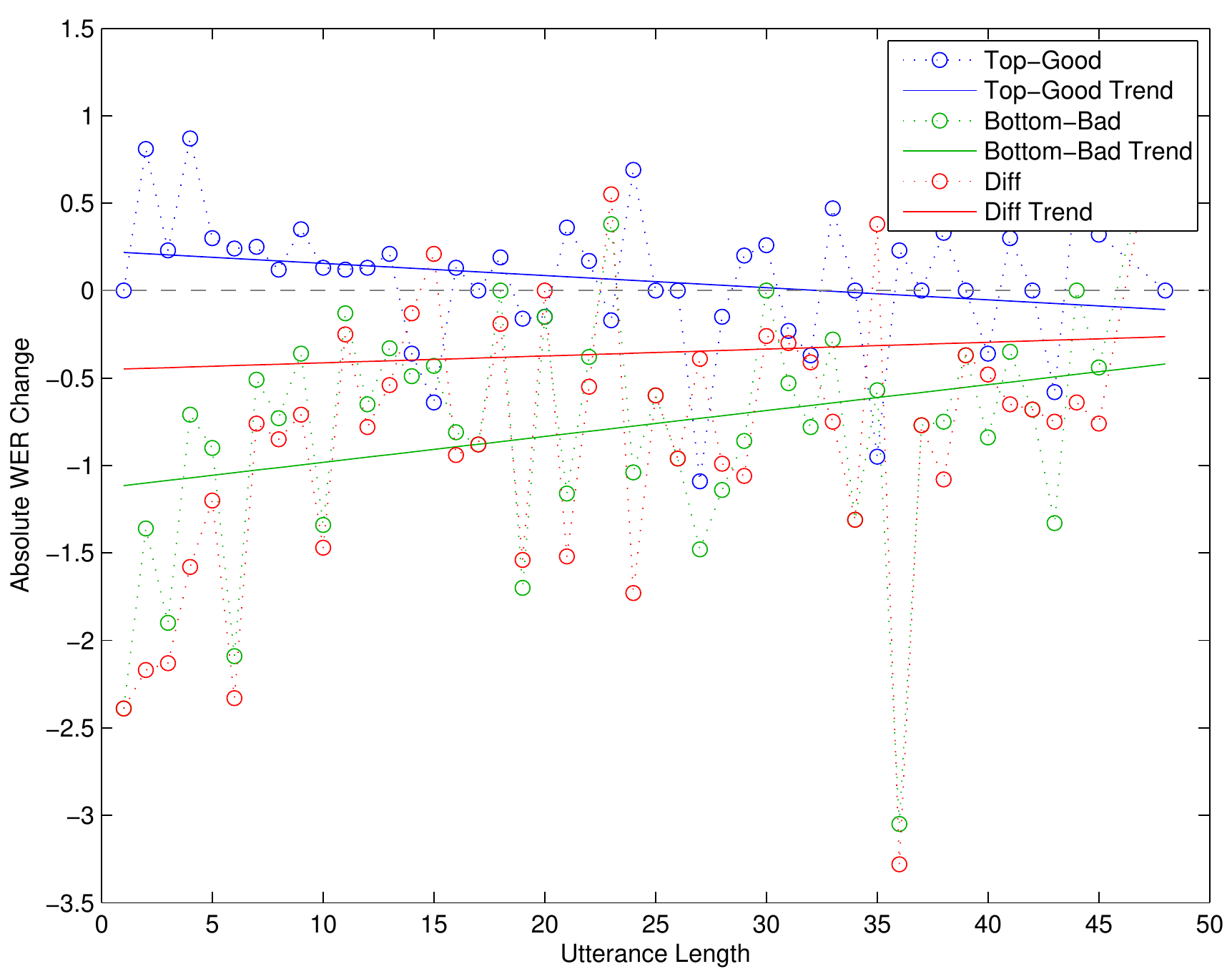}
    \caption{Test: NCPCM + 5gram NNLM + MERT(W)}
		\label{fig:5gNNLM_test}
  \end{subfigure}
	\begin{subfigure}[t]{0.47\textwidth}
		\centering
		\includegraphics[width=\textwidth]{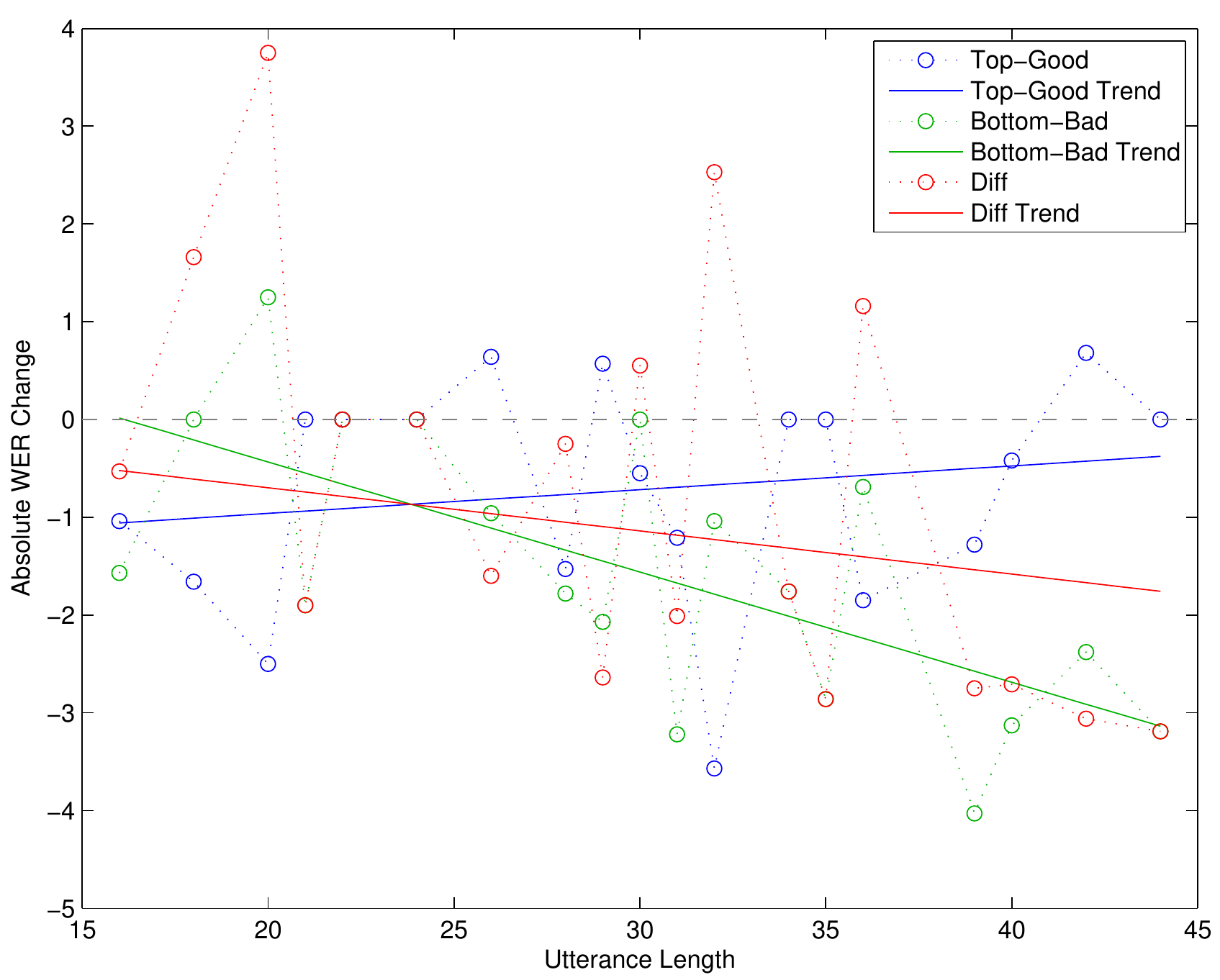}
	  \caption{Out-of-Domain Dev: NCPCM + generic LM + MERT(W)}
		\label{fig:out_of_domain_dev}
	\end{subfigure}
  \hfill
  \begin{subfigure}[t]{0.47\textwidth}
	  \centering
		\includegraphics[width=\textwidth]{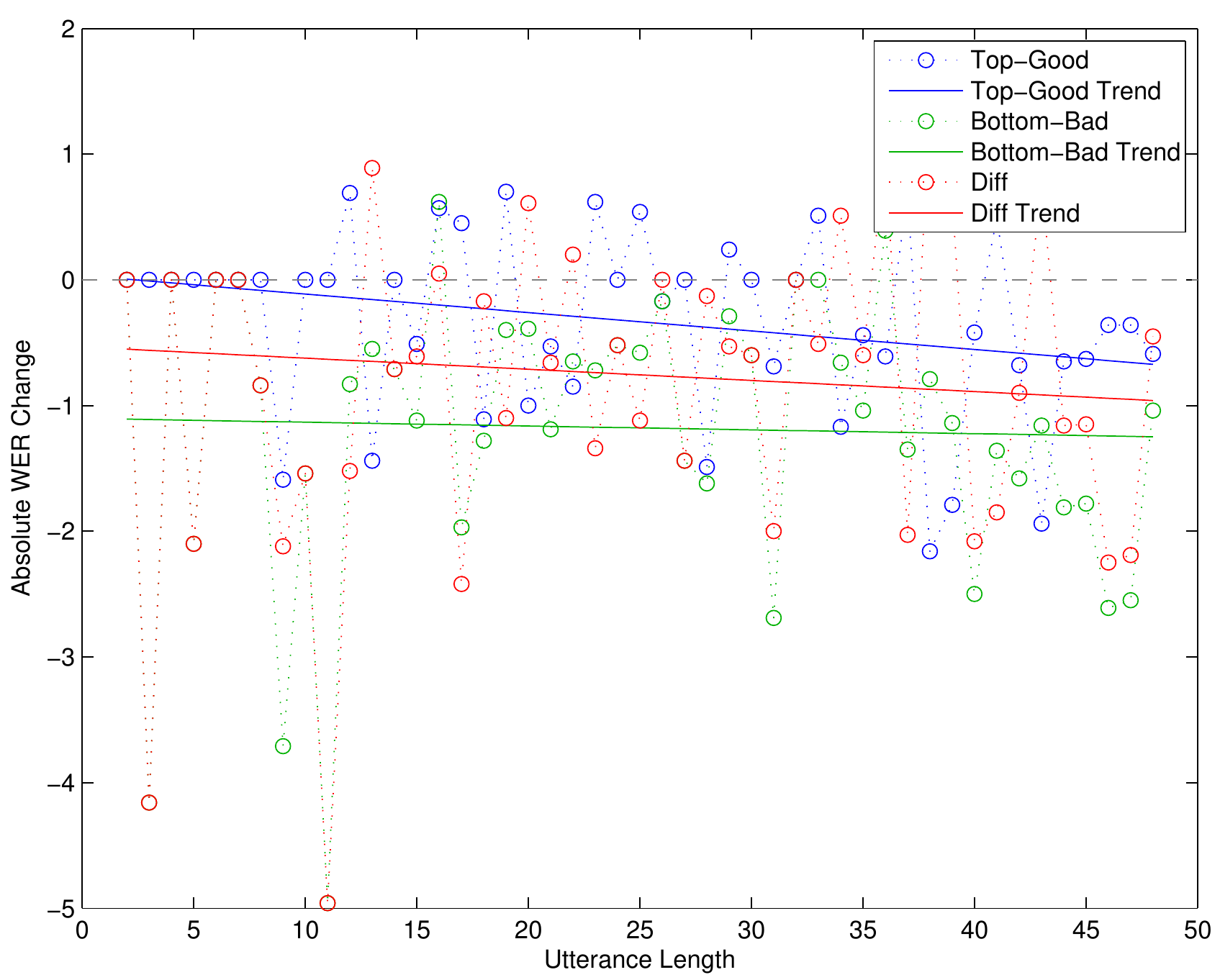}
    \caption{Out-of-Domain Test: NCPCM + generic LM + MERT(W) }
		\label{fig:out_of_domain_test}
	\end{subfigure}
	\caption{Length of hypotheses through our NCPCM models versus absolute WER change.\\
    Blue \& Green lines represent difference between WER of our system and the baseline ASR, for top-good and bottom-bad hypotheses, respectively.      In an ideal scenario, all these lines would be below 0, thus all providing a change in WER towards improving the system. However we see in some cases that the WER increases, especially when the hypotheses length is short and when the performance is good. This is as expected since from Fig.~\ref{fig:domain_split} some cases are at 0\% WER  due to the already highly-optimized nature of our ASR.\\
    The red line represents the aggregate error over all data for each word length and as we can see in \textbf{all} cases the trend is one of \textbf{improving} the WER.
    This matches {\color{red}Hypotheses D, E, F, G}
  \label{fig:wer_analysis}}
\end{figure*}

\subsection{Better Recovery during Poor Recognitions}\label{res:low_conditions}
To justify the claim that our system can offset for the performance deficit of the ASR at tougher conditions (as per \emph{Hypothesis}~\ref{hyp-D}), we formulate a sub-problem as follows:

\vspace{2mm}\noindent\textbf{Problem Formulation:} We divide equally, per sentence length, our development and test datasets into good recognition results (top-good) and poor recognition results (bottom-bad) subsets based on the WER of the ASR and analyze the improvements and any degradation caused by our system.

Figure~\ref{fig:wer_analysis} shows the plots of the above mentioned analysis for different systems as captioned. The blue lines are representative of the improvements provided by our system for top-good subset over different utterance lengths, i.e., it indicates the difference between our system and the original WER of the ASR (negative values indicate improvement and positive values indicate degradation resulting from our system). The green lines indicate the same for bottom-bad subset of the database. The red indicates the difference between the bottom-bad WERs and the top-good WERs, i.e., negative values of red indicate that the system provides more improvements to the bottom-bad subset relative to the top-good subset. The solid lines represent their respective trends which is obtained by a simple linear regression (line-fitting).

For poor recognitions, we are concerned about the bottom-bad subset, i.e., the green lines in Figure~\ref{fig:wer_analysis}. Firstly, we see that the solid green line is always below zero, which indicates there is always improvements for bottom-bad i.e., poor recognition results. Second, we observe that the solid red line usually stays below zero, indicating that the performance gains made by the system add more for the bottom-bad poor recognition results compared to the top-good subset (good recognitions). Further, more justifications are provided later in the context of out-of-domain task (Section~\ref{sec:res}~\ref{res:ood}) where high mismatch results in tougher recognition task are discussed.

\begin{table*}[t]
    \centering
    \begin{tabular}{|c|c|c|c|c|}
        \hline
      \multicolumn{5}{|c|}{In domain testing on Fisher Data}  \\
        \hline
        \multirow{2}{*}{Method} & \multicolumn{2}{c|}{Dev} & \multicolumn{2}{c|}{Test} \\
         & WER & BLEU & WER & BLEU \\
         \hline
         ASR output (Baseline-1) & 15.46\% & 75.71 & 17.41\% & 72.99 \\
				 \hline
				 ASR + RNNLM re-scoring (Baseline-2) & 16.17\% & 74.39 & 18.39\% & 71.24 \\
				 \hline
				 Word based + bigram LM (Baseline-3) & 16.23\% & 74.28 & 18.10\% & 71.76 \\
				 \hline
				 Word based + bigram LM + MERT(B) & 15.46\% & 75.70 & \textbf{17.40\%} & 72.99 \\
				 
				 Word based + bigram LM + MERT(W) & \textbf{15.39\%} & 75.65 & \textbf{17.40\%} & 72.77 \\
				 \hline
				 Word based + trigram LM + MERT(B) & 15.48\% & 75.59 & 17.47\% & 72.81 \\
				 
				 Word based + trigram LM + MERT(W) & 15.46\% & 75.46 & 17.52\% & 72.46 \\
				 \hline
				 DLM (Baseline-4) & 23.65\% & 63.35 & 25.36\% & 61.19 \\
				 DLM w/ extended feats & 24.48\% & 62.92 & 26.12\% & 60.98 \\
				 \hline
				 \hline
				 \hline
         Proposed NCPCM & 20.33\% & 66.70 & 22.32\% & 63.81 \\
         \hline
         NCPCM + MERT(B) & \textbf{15.11\%} & \textbf{76.06} & \textbf{17.18\%} & \textbf{73.00} \\
         \hline
         NCPCM + MERT(W) & \textcolor{red}{\textbf{15.10\%}} & \textbf{76.08} & \textbf{17.15\%} & \textbf{73.05} \\
         \hline
         NCPCM + MERT(B) w/o re-ordering & \textbf{15.27\%} & \textbf{76.02} & \textcolor{red}{\textbf{17.11\%}} & \textcolor{red}{\textbf{73.33}} \\
         \hline
         NCPCM + MERT(W) w/o re-ordering & \textbf{15.19\%} & \textbf{75.90} & \textbf{17.18\%} & \textbf{73.04} \\
         \hline
         NCPCM + 10best + MERT(B) & \textbf{15.19\%} & \textcolor{red}{\textbf{76.12}} & \textbf{17.17\%} & \textbf{73.22} \\
         \hline
         NCPCM + 10best + MERT(W) & \textbf{15.16\%} & \textbf{75.91} & \textbf{17.21\%} & \textbf{73.03} \\
         \hline
    \end{tabular}
		\caption{Noisy-Clean Phrase Context Model (NCPCM) results (uses exactly same LM as ASR)}
    \label{tab:results}
\end{table*}

\subsection{Improvements under all Acoustic Conditions}\label{res:wer}
To justify the claim that our system can consistently provide benefits over any ASR system (\emph{Hypothesis}~\ref{hyp-E}), we need to show that the proposed system: (i) does not degrade the performance of the good recognition, (ii) provides improvements to poor recognition instances, of the ASR. The latter has been discussed and confirmed in the previous Section~\ref{sec:res}~\ref{res:low_conditions}. For the former, we provide evaluations from two point of views: (1) assessment of WER trends of top-good and bottom-bad subsets (as in the previous Section~\ref{sec:res}~\ref{res:low_conditions}), and (2) overall absolute WER of the proposed systems.

Firstly, examining Figure~\ref{fig:wer_analysis}, we are mainly concerned about the top-good subset pertaining to degradation/improvement of good recognition instances. We observe that the solid blue line is close to zero in all the cases, which implies that the degradation of good recognition is extremely minimal. Moreover, we observe that the slope of the line is almost zero in all the cases, which indicates that the degradation is minimal and mostly consistent over different utterance lengths. Moreover, assessing the degradation from the absolute WER perspective, Figure~\ref{fig:in_domain} shows the WER over utterance lengths for the top-good and bottom-bad subsets for the in-domain case. The top-good WER is small, at times even 0\% (perfect recognition) thereby allowing very small margin for improvement. In such a case, we see minimal degradation. Although we lose a bit on very good recognitions which is extremely minimal, we gain significantly in the case of `bad' recognitions. Thus to summarize, the damage that this system can make, under the best ASR conditions, is minimal and offset by the potential significant gains present when the ASR hits some tough recognition conditions.

\vspace{2mm}
\noindent\textbf{WER experiments:}\\
\indent Secondly, examining the overall WER, Table~\ref{tab:results} gives the results of the baseline systems and the proposed technique. Note that we use the same language model as the ASR. This helps us evaluate a system that does not include additional information. We provide the performance measures on both the development and held out test data. The development data is used for MERT tuning. 
\vspace{2mm}\linebreak\noindent\textbf{Baseline results:} 
The output of the ASR (Baseline-1) suggests that the development data is less complex compared to the held out test set.
In our case, the RNN-LM based lattice re-scoring (Baseline-2) doesn't help.
This results shows that even with a higher order context, the RNN-LM is unable to recover the errors present in the lattice, suggesting that the errors stem from pruning during decoding.
We note that the word-based system (Baseline-3) doesn't provide any improvements.
Even when we increase context (trigram LM) and use MERT optimization, the performance is just on par with the original ASR output. 
Further, DLM re-ranking (Baseline-4) fails to provide any improvements in our case. 
This result is in conjunction with the finding in \cite{bikelf2012confusion}, where the DLM provides improvements only when used in combination with ASR baseline scores.
However, we believe introduction of ASR scores into NCPCM can be beneficial as would be in the case of DLMs.
Thus, to demonstrate the independent contribution of NCPCM vs DLM's, rather than investigate fusion methods, we don't utilize baseline ASR scores for either of the two methods.
We plan to investigate the benefits of multi-method fusion in our future work.
When using the extended feature set for training the DLM, we don't observe improvements.
With our setup, none of the baseline systems provide noticeable significant improvements over the ASR output.
We believe this is due to the highly optimized ASR setup, and the nature of the database itself being noisy telephone conversational speech.
Overall, the results of baseline highlights: (i) the difficulty of the problem for our setup, (ii) re-scoring is insufficient and emphasizes the need for recovering pruned out words in the output lattice.
\vspace{2mm}\linebreak\noindent\textbf{NCPCM results:}
The NCPCM is an ensemble of phrase translation model, language model, word penalty model and re-ordering models.
Thus the tuning of the weights associated with each model is crucial in the case of the phrase based models \cite{neubig2016optimization}.
The NCPCM without tuning, i.e., assigning random weights to the various models, performs very poorly as expected.
The word-based model lacks re-ordering/distortion modeling and word penalty models and hence are less sensitive to weight tuning. 
Thus it is unfair to compare the un-tuned phrase based models with the baseline or word-based counterpart.
Hence, for all our subsequent experiments, we only include results with MERT.
When employing MERT, all of the proposed NCPCM systems significantly outperform the baseline (statistically significant with $p < 0.001$ for both word error and sentence error rates \cite{gillick1989some} with 51,230 word tokens and 4,914 sentences as part of the test data).
We find that MERT optimized for WER consistently outperforms that with optimization criteria of BLEU score. We also perform trials by disabling the distortion modeling and see that results remain relatively unchanged.  This is as expected since the ASR preserves the sequence of words with respect to the audio and there is no reordering effect over the errors. The phrase based context modeling provides a relative improvement of 1.72\% (See Table~\ref{tab:results}) over the baseline-3 and the ASR output.
Using multiple hypotheses (10-best) from the ASR, we hope to capture more relevant error patterns of the ASR model, thereby enriching the noisy channel modeling capabilities. However, we find that the 10-best gives about the same performance as the 1-best. In this case we considered 10 best as 10 separate training pairs for training the system. In the future we want to exploit the inter-dependency of this ambiguity (the fact that all the 10-best hypotheses represent a single utterance) for training and error correction at test time.

\begin{table*}[t]
    \centering
    \begin{tabular}{|c|c|c|c|c|c|c|}
        \hline
      \multicolumn{7}{|c|}{Cross domain testing on TED-LIUM Data}  \\
        \hline
        \multirow{2}{*}{Method} & \multicolumn{2}{|c|}{Dev} & \multicolumn{4}{|c|}{Test} \\
         & WER & BLEU & WER & $\Delta_1$ & $\Delta_2$ & BLEU \\
         \hline
						Baseline-1 (ASR) & 26.92\% & 62.00 & 23.04\% & 0\% & -10.9\% & 65.71 \\
						
						ASR + RNNLM re-scoring (Baseline-2) & \textbf{24.05\%} & \textbf{64.74} & \textbf{20.78\%} & \textbf{9.8\%} & 0\% & \textbf{67.93} \\
						
						Baseline-3 (Word-based) & 29.86\% & 57.55 & 25.51\% & -10.7\% & -22.8\% & 61.79 \\
						
						Baseline-4 (DLM) & 33.34\% & 53.12 & 28.02\% & -21.6\% & -34.8\% & 58.50 \\
						DLM w/ extended feats & 30.51\% & 57.14 & 29.33\% & -27.3\% & -41.1\% & 57.60 \\
						\hline
						NCPCM + MERT(B) & \textbf{26.06\%} & \textbf{63.30} & \textbf{22.51\%} & \textbf{2.3\%} & -8.3\% & \textbf{66.67} \\
						
		 				NCPCM + MERT(W) & \textbf{26.15\%} & \textbf{63.10} & \textbf{22.74\%} & \textbf{1.3\%} & -9.4\% & \textbf{66.36} \\
						
						\hline
						NCPCM + generic LM + MERT(B) & \textbf{25.57\%} & \textbf{63.98} & \textbf{22.38\%} & \textbf{2.9\%} & -7.7\% & \textbf{66.97} \\

						NCPCM + generic LM + MERT(W) & \textbf{25.56\%} & \textbf{63.83} & \textbf{22.33\%} & \textbf{3.1\%} & -7.5\% & \textbf{66.96} \\
						\hline
						RNNLM re-scoring + NCPCM + MERT(B) & \textbf{23.36\%} & \textbf{65.88} & \textbf{20.40\%} & \textbf{11.5\%} & \textbf{1.8\%} & \textbf{68.39} \\
						RNNLM re-scoring + NCPCM + MERT(W) & \textbf{23.32\%} & \textbf{65.76} & \textbf{20.57} & \textbf{10.7\%} & \textbf{1\%} & \textbf{68.07} \\
						\hline
						RNNLM re-scoring + NCPCM + generic LM + MERT(B) & \textbf{23.00\%} & \textcolor{red}{\textbf{66.48}} & \textbf{20.31\%} & \textbf{11.8\%} & \textbf{2.3\%} & \textcolor{red}{\textbf{68.52}} \\
						RNNLM re-scoring + NCPCM + generic LM + MERT(W) & \textcolor{red}{\textbf{22.80\%}} & \textbf{66.19} & \textcolor{red}{\textbf{20.23\%}} & \textcolor{red}{\textbf{12.2\%}} & \textcolor{red}{\textbf{2.6\%}} & \textbf{68.49} \\
						\hline
    \end{tabular}
		\captionsetup{justification=centering}
    \caption{Results for out-of-domain adaptation using Noisy-Clean Phrase Context Models (NCPCM)\\
		$\Delta_1$:Relative \% improvement w.r.t baseline-1; $\Delta_2$:Relative \% improvement w.r.t baseline-2;}
    \label{tab:ood}
\end{table*}

\subsection{Adaptation}\label{res:ood}
\noindent\textbf{WER experiments:}
\par To assess the adaptation capabilities, we evaluate the performance of the proposed noisy-clean phrase context model on an out-of-domain task, TED-LIUM data-base, shown in Table~\ref{tab:ood}.

\vspace{2mm}\noindent\textbf{Baseline Results:}
The baseline-1 (ASR performance) confirms of the heightened mismatched conditions between the training Fisher Corpus and the TED-LIUM data-base.
Unlike in matched in-domain evaluation, the RNNLM re-scoring provides drastic improvements (9.8\% relative improvement with WER) when tuned with out-of-domain development data set.
The mismatch in cross domain evaluation reflects in considerably worse performance for the word-based and DLM baselines (compared to matched conditions).

\vspace{2mm}\noindent\textbf{NCPCM Results:}
However, we see that the phrase context modeling provides modest improvements over the baseline-1 of approximately 2.3\% (See Table~\ref{tab:ood}) relative on the held-out test set. We note that the improvements are consistent compared to the earlier in-domain experiments in Table~\ref{tab:results}. Moreover, since the previous LM was trained on Fisher Corpus, we adopt a more generic English LM which provides further improvements of up to 3.1\% (See Table~\ref{tab:ood}).
 
We also experiment with NCPCM over the re-scored RNNLM output.
We find the NCPCM to always yield consistent improvements over the RNNLM output (See $\Delta_1$ \& $\Delta_2$ in Table~\ref{tab:ood}).
An overall gains of 2.6\% relative is obtained over the RNNLM re-scored output (baseline-2) i.e., 12.2\% over ASR (baseline-1) is observed.
This confirms that the NCPCM is able to provide improvements parallel, in conjunction to the RNNLM or any other system that may improve ASR performance and therefore supports the \emph{Hypothesis}~\ref{hyp-E} in yielding improvements in the highly optimized ASR environments.
This also confirms the robustness of the proposed approach and its application to the out-of-domain data. More importantly, the result confirms \emph{Hypothesis}~\ref{hyp-F}, i.e., our claim of rapid adaptability of the system to varying mismatched acoustic and linguistic conditions.
The extreme mismatched conditions involved in our experiments supports the possibility of going one step further and training our system on artificially generated data of noisy transformations of phrases as in \cite{tan2010automatic,sagae2012hallucinated,celebi2012semi,kurata2011training,dikici2012performance,xu2012phrasal}.
Thus possibly eliminating the need for an ASR for training purposes.

Further, comparing the WER trends from the in-domain task (Figure~\ref{fig:1best_test}) to the out-of-domain task (Figure~\ref{fig:out_of_domain_test}), we firstly find that the improvements in the out-of-domain task are obtained for both top-good (good recognition) and bottom-bad (bad recognition), i.e., both the solid blue line and the solid green line are always below zero. Secondly, we observe that the improvements are more consistent throughout all the utterance lengths, i.e., all the lines have near zero slopes compared to the in-domain task results.
Third, comparing Figure~\ref{fig:in_domain} with Figure~\ref{fig:out_of_domain}, we observe more room for improvement, both for top-good portion as well as the bottom-bad WER subset of data set. 
The three findings are fairly meaningful considering the high mismatch of the out-of-domain data.

\begin{table*}[t]
    \centering
    \begin{tabular}{|c|c|c|c|c|}
        \hline
      \multicolumn{5}{|c|}{In domain testing on Fisher Data}  \\
        \hline
        \multirow{2}{*}{Method} & \multicolumn{2}{|c|}{Dev} & \multicolumn{2}{|c|}{Test} \\
         & WER & BLEU & WER & BLEU \\
         \hline
         Baseline-1 (ASR output) & 15.46\% & 75.71 & 17.41\% & 72.99 \\
         \hline
				 Baseline-2 (ASR + RNNLM re-scoring) & 16.17\% & 74.39 & 18.39\% & 71.24 \\
				 \hline
				 Baseline-3 (Word based + 5gram NNLM) & 15.47\% & 75.63 & 17.41\% & 72.92 \\
				 Word based + 5gram NNLM + MERT(B) & 15.46\% & 75.69 & \textbf{17.40\%} & 72.99 \\
				 Word based + 5gram NNLM + MERT(W) & \textbf{15.42\%} & 75.58 & \textbf{17.38\%} & 72.75 \\
				 \hline
						NCPCM + 3gram NNLM + MERT(B) & \textbf{15.46\%} & \textbf{75.91} & \textbf{17.37\%} & \textbf{73.24} \\
										 \hline
						NCPCM + 3gram NNLM + MERT(W) & \textbf{15.28\%} & \textbf{75.94} & \textbf{17.11\%} & \textbf{73.31} \\
						\hline
										 
						NCPCM + 5gram NNLM + MERT(B) & \textbf{15.35\%} & \textcolor{red}{\textbf{75.99}} & \textbf{17.20\%} & \textcolor{red}{\textbf{73.34}} \\
						\hline
						NCPCM + 5gram NNLM + MERT(W) & \textcolor{red}{\textbf{15.20\%}} & \textbf{75.96} & \textcolor{red}{\textbf{17.08\%}} & \textbf{73.25} \\
						\hline
										 
						NCPCM + NNJM-LM (5,4) + MERT(B) & \textbf{15.29\%} & \textbf{75.93} & \textbf{17.13\%} & \textbf{73.26} \\
						\hline
						NCPCM + NNJM-LM (5,4) + MERT(W) & \textbf{15.28\%} & \textbf{75.94} & \textbf{17.13\%} & \textbf{73.29} \\
						\hline
    \end{tabular}
    \caption{Results for Noisy-Clean Phrase Context Models (NCPCM) with Neural Network Language Models (NNLM) and Neural Network Joint Models (NNJM)}
    \label{tab:lm_results}
		\vspace{-3mm}
\end{table*}

\subsection{Exploit Longer Context}\label{res:long_context}
Firstly, inspecting the error correction results from Table~\ref{tab:analysis}, cases 2 and 4 hint at the ability of the system to select appropriate word-suffixes using long term context information.

Second, from detailed WER analysis in Figure~\ref{fig:wer_analysis}, we see that the bottom-bad (solid green line) improvements decrease with increase in length in most cases, hinting at potential improvements to be found by using higher contextual information for error correction system as future research directions. Moreover, closer inspection across different models, comparing the trigram MLE model (Figure~\ref{fig:1best_test}) with the 5gram NNLM (Figure~\ref{fig:5gNNLM_test}), we find that the NNLM provides minimal degradation and better improvements especially for longer utterances by exploiting more context (the blue solid line for NNLM has smaller intercept value as well as higher negative slope). We also find that for the bottom-bad poor recognition results (green solid-line), the NNLM gives consistent (smaller positive slope) and better improvements especially for the higher length utterances (smaller intercept value). Thus emphasizing the gains provided by higher context NNLM.

\vspace{2mm}\noindent\textbf{WER experiments:}
Third, Table~\ref{tab:lm_results} shows the results obtained using a neural network language model of higher orders (also trained only on the in-domain data).
For a fair comparison, we adopt a higher order (5gram) NNLM for the baseline-3 word based noise channel modeling system.
Even with a higher order NNLM, the baseline-3 fails to improve upon the ASR.
We don't include the baseline-4 results under this section, since DLM doesn't include a neural network model.
Comparing results from Table~\ref{tab:results} with Table~\ref{tab:lm_results}, we note the benefits of higher order LMs, with the 5-gram neural network language model giving the best results (a relative improvement of 1.9\% over the baseline-1), outperforming the earlier MLE n-gram models as per \emph{Hypothesis}~\ref{hyp-G}. 

Moreover, experimental comparisons with baseline-3 (word-based) and NCPCM models, both incorporating identical 5-gram neural network language models confirms the advantages of NCPCM (a relative improvement of 1.7\%). However, the neural network joint model LM with target context of 5 and source context of 4 did not show significant improvements over the traditional neural LMs. We expect the neural network models to provide further improvements with more training data.

\subsection{Regularization}\label{res:regularization}
Finally, the last case in Table~\ref{tab:analysis} is of text regularization as described in Section~\ref{sec:hyp}, \emph{Hypothesis}~\ref{hyp-H}. Overall, in our experiments, we found that approximately 20\% were cases of text regularization and the rest were a case of the former hypotheses.

\vspace{-1mm}
\section{Conclusions \& Future Work}\label{sec:conclusion}
In this work, we proposed a noisy channel model for error correction based on phrases. The system post-processes the output of an automated speech recognition system and as such any contributions in improving ASR are in conjunction of NCPCM. 
We presented and validated a range of hypotheses.
Later on, we supported our claims with apt problem formulation and their respective results.
We showed that our system can improve the performance of the ASR by
(i) re-scoring the lattices (\emph{Hypothesis}~\ref{hyp-A}),
(ii recovering words pruned from the lattices (\emph{Hypothesis}~\ref{hyp-B}),
(iii) recovering words never seen in the vocabulary and training data (\emph{Hypothesis}~\ref{hyp-C}),
(iv) exploiting longer context information (\emph{Hypothesis}~\ref{hyp-G}), and
(v) by regularization of language syntax (\emph{Hypothesis}~\ref{hyp-H}).
Moreover, we also claimed and justified that our system can provide more improvement in low-performing ASR cases (\emph{Hypothesis}~\ref{hyp-D}), while keeping the degradation to minimum in cases when the ASR performs well (\emph{Hypothesis}~\ref{hyp-E}).
In doing so, our system could effectively adapt (\emph{Hypothesis}~\ref{hyp-F}) to changing recognition environments and provide improvements over any ASR systems.


In our future work, the output of the noisy-clean phrase context model will be fused with the ASR beliefs to obtain a new hypothesis. We also intend to introduce ASR confidence scores and signal SNR estimates, to improve the channel model. We are investigating introducing the probabilistic ambiguity of the ASR in the form of lattice or confusion networks as inputs to the channel-inversion model.

Further, we will utilize sequence-to-sequence (Seq2seq) translation modeling \cite{sutskever2014sequence} to map ASR outputs to reference transcripts. The Seq2seq model has been shown to have benefits especially in cases where training sequences are of variable length \cite{cho2014properties}. We intend to employ Seq2seq model to encode ASR output to a fixed-size embedding and decode this embedding to generate the corrected transcripts.
\vspace{-2mm}
\section*{Financial Support}
\vspace{-1mm}
The U.S. Army Medical Research Acquisition Activity, 820 Chandler Street, Fort Detrick MD 21702- 5014 is the awarding and administering acquisition office. This work was supported by the Office of the Assistant Secretary of Defense for Health Affairs through the Psychological Health and Traumatic Brain Injury Research Program under Award No. W81XWH-15-1-0632. Opinions, interpretations, conclusions and recommendations are those of the author and are not necessarily endorsed by the Department of Defense.

\bibliographystyle{IEEEbib}
\bibliography{mybib}

\end{document}